\documentclass[10pt,twocolumn,letterpaper]{article}
\usepackage{cvpr}              
\usepackage[utf8]{inputenc}
\usepackage{times}
\usepackage{epsfig}
\usepackage{graphicx}
\usepackage{amsmath}
\usepackage{amssymb}
\usepackage{booktabs}
\usepackage{multirow}
\usepackage{tabularx}
\usepackage{hyperref}

\title{Deep Semantic Iterative Reconstruction:  One-Shot Universal Medical Anomaly Detection}

\author{
NingZhu\\
}

\begin{document}

\maketitle

\begin{abstract}
Unsupervised anomaly detection in medical imaging is severely limited by the scarcity of normal samples and poor cross-modality generalization. Existing methods typically train dedicated models per dataset or modality, requiring hundreds of normal images and struggling to generalize across imaging protocols.

We present Deep Semantic Iterative Reconstruction (DSIR), a simple yet powerful convolutional distillation framework that trains a single universal model for robust anomaly detection across heterogeneous medical domains using extremely limited normal data. DSIR employs a frozen pretrained teacher encoder to extract multi-scale deep semantic features and a simple re-downsampling decoder that performs iterative semantic reconstruction via multiple refinement loops in a single forward pass. By mixing exactly one normal sample from each of nine diverse datasets, it learns strong cross-domain normality priors in deep feature space and generalizes effectively to unseen domains without task-specific retraining.

Extensive experiments on nine challenging benchmarks spanning chest X-rays, retinal fundus, skin lesions, and brain MRI show that DSIR achieves state-of-the-art performance across all four protocols---one-shot universal, full-shot universal, one-shot specialized, and full-shot specialized---with notable gains on difficult datasets.
\end{abstract}

\section{Introduction}
Unsupervised medical anomaly detection enables the identification of abnormalities across diverse imaging modalities such as chest X-rays, retinal fundus images, skin lesions, and brain MRI scans \cite{rsna,oct,aptos,isic,brat}. However, this task faces two primary challenges. First, the scarcity of normal samples, the rarity of pathologies, and the high cost of annotation severely limit data availability. Second, achieving effective generalization with a single model across different imaging modalities, disease types, and acquisition protocols remains difficult.

Most existing unsupervised anomaly detection approaches \cite{zhou2017anomaly,kingma2013auto,bergmann2019improving,kascenas2022denoising,gong2019memorizing,schlegl2019fanogan,akcay2019ganomaly,beizaee2025correcting,bercea2023mask,liu2025survey,jin2025dual} require a large number of normal samples from a single dataset to train a dedicated model. These methods deliver strong performance within their target domain but fail to generalize across modalities or diseases \cite{huang2024adapting,luo2025exploring,zhu2024toward,gu2024anomalygpt,cai2025medianomaly}. Consequently, constructing separate models for each clinical task is resource-intensive and impractical in real-world scenarios where normal samples are limited and cross-task data sharing is restricted.

To overcome these limitations, we propose DSIR, a  framework that trains a single universal model capable of detecting anomalies across diverse medical domains using few normal samples. The key innovation is to construct a small but highly heterogeneous training set by selecting exactly one normal sample from each of nine representative medical datasets, resulting in a total of only nine images. This per-dataset one-shot sampling strategy enables strong cross-modality generalization without requiring task-specific retraining.

DSIR extends the reverse distillation paradigm by employing a pretrained teacher encoder to extract multi-scale deep features and a simple student decoder equipped with multiple iterative refinement loops within a single forward-backward pass. These loops progressively enforce robust normality priors on the deep representations. During inference, anomaly scores are computed through multi-scale cosine similarity aggregation followed by Gaussian smoothing and cross-loop fusion.

Our main contributions are as follows:
\begin{itemize}
\item We propose a semantic reconstruction framework for one-shot universal medical anomaly detection. 

\item We introduce a multi-loop iterative refinement mechanism that strengthens normality constraints in deep feature space, thereby enabling robust generalization and strong universality across diverse medical modalities with limited normal samples.

\item Extensive experiments on nine diverse medical benchmarks demonstrate that our method achieves state-of-the-art performance across one-shot universal, full-shot universal, one-shot specialized, and full-shot specialized settings.
\end{itemize}

\begin{figure*}[t]
\centering
\includegraphics[width=\textwidth]{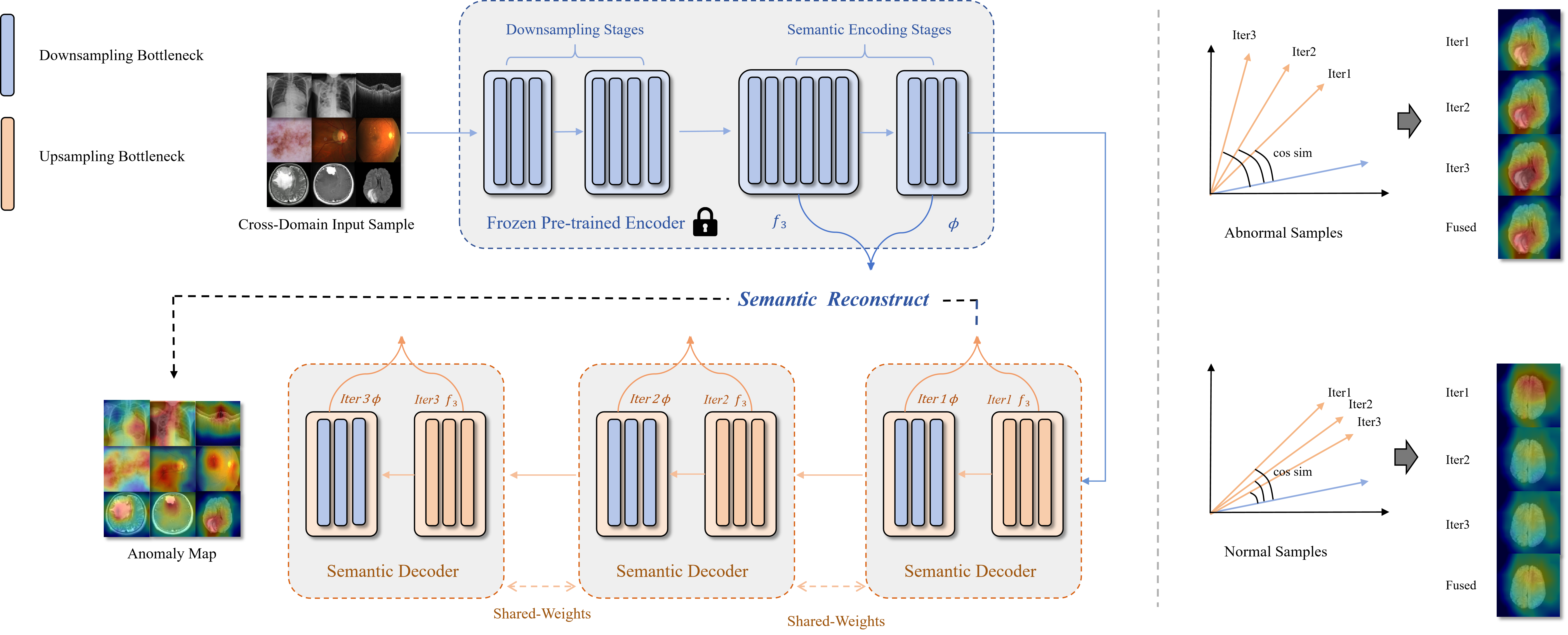}
\caption{Overview of the DSIR framework. The architecture consists of a frozen pre-trained teacher encoder \(E\) and student decoder \(D\). During training, the teacher encoder processes one normal image from each of multiple source domains, extracting multi-scale deep features. The student decoder then performs iterative semantic reconstruction on the most compressed features through recurrent loops. At inference time, for any previously unseen target domain, the same single model receives only a test image \(x\), and the anomaly score and pixel-level anomaly map are finally obtained from the reconstruction discrepancies accumulated across all iterative loops.}
\label{fig:overview}
\end{figure*}

\section{Related Work}
Unsupervised anomaly detection has been extensively studied in computer vision, with methods broadly categorized into reconstruction-based, memory-bank-based, and knowledge distillation-based approaches.

\subsection{Full-shot Anomaly Detection}
Traditional full-shot anomaly detection methods rely on large amounts of normal training samples from a single domain and can be broadly divided into pixel-reconstruction and feature-reconstruction approaches.

Pixel-level reconstruction methods typically employ generative models to directly reconstruct the input image, aiming to produce small reconstruction residuals for normal samples and large errors for anomalous samples, thereby enabling anomaly localization through pixel- or structure-level differences. Early works primarily adopt AutoEncoder (AE) and Variational AutoEncoder (VAE) as representatives \cite{zhou2017anomaly,kingma2013auto,bergmann2019improving,kascenas2022denoising,gong2019memorizing,meissen2023unsupervised}, learning the normal data distribution by minimizing pixel-wise reconstruction error. Subsequent GAN-based methods were explored as alternatives to improve reconstruction quality\cite{schlegl2019fanogan,akcay2019ganomaly}. Recent trends have shifted toward Diffusion models for generating high-quality normal samples \cite{beizaee2025correcting,bercea2023mask,liu2025survey,jin2025dual}. In addition, Transformer architectures have also been explored to construct these generative models \cite{nafez2025patchguard}, enhancing the model’s ability to capture global contextual information.

Early studies primarily employ Feature AutoEncoders (FAEs) \cite{wang2021student,guo2023recontrast} that perform reconstruction on multi-layer feature maps extracted from pre-trained backbones, minimizing feature reconstruction error to capture the normal data manifold. The introduction of Teacher-Student Distillation frameworks, pioneered by RD4AD \cite{deng2022anomaly}, has since established the dominant paradigm for feature-level reconstruction. Building upon this foundation, a substantial body of subsequent work\cite{tien2023revisiting,guo2025dinomaly,you2023adtr,guo2024encoder,tang2025anomaly} has extended and refined the approach, yielding significant performance gains in both medical and industrial anomaly detection scenarios.

However, these full-shot methods typically require hundreds to thousands of normal samples per dataset and train domain-specific models, limiting their applicability in data-scarce medical scenarios where normal samples are extremely limited.

\subsection{Few-shot Anomaly Detection}
To address the scarcity of normal training samples, recent studies have explored few-shot and one-shot anomaly detection. These approaches can be grouped into adaptation-based, synthesis-based, and distillation-based categories.

Adaptation-based methods transfer techniques from few-shot classification to anomaly detection. Some leverage pre-trained foundation models such as DINOv2 \cite{damm2025anomalydino} for patch-based few-shot adaptation. InCTRL\cite{zhu2024toward} employ in-context residual learning with few-shot sample prompts or INPFormer\cite{luo2025exploring} explore intrinsic normal prototypes within a single image to achieve universal anomaly detection. In medical imaging, MVFA-AD\cite{huang2024adapting} employ visual-language models adapted for generalizable anomaly detection across diverse modalities. AnomalyGPT\cite{gu2024anomalygpt} further advances this direction by employing large vision-language models with prompt engineering and in-context learning to perform direct anomaly classification and localization in industrial settings.

Synthesis-based methods\cite{gui2025few,li2024one} generate synthetic anomalies to augment training data. Few-shot anomaly-driven generation synthesizes realistic lesions from one or few normal samples to enhance anomaly diversity and improve classification and segmentation performance.

Distillation-based methods apply knowledge distillation in few-shot settings. D24FAD\cite{dong2026dual} employs dual-path teacher-student and reverse distillation constraints to enhance normal feature learning and anomaly localization under limited data.

\section{Methodology}

\subsection{Problem Definition}
During training, we are given a collection of source domains \(\{ \mathcal{D}_i \}_{i=1}^N\). For each domain \(\mathcal{D}_i\), we are provided with exactly one normal image \(x_i \in \mathcal{D}_i\). All such single-sample images are combined into one unified training set \(\mathcal{X}_{\text{train}} = \{ x_i \}_{i=1}^N\), and a single model is trained once on this extremely limited collection.

At inference time, the same model takes a single test image \(x\) from any previously unseen target domain and directly outputs an anomaly score \(s(x)\) and a pixel-level anomaly map \(M(x)\).

\subsection{Semantic Reconstruction Framework}
The overall architecture consists of a frozen pretrained teacher encoder and a  student decoder. The teacher extracts two hierarchical deep feature maps: the mid-level semantic feature after its third residual stage and the most compressed high-level semantic feature after the fourth stage.

The student decoder takes only the most compressed teacher feature as input. It first employs a transposed convolution to upsample and produce an intermediate representation that aligns with the teacher’s mid-level semantics. Subsequently, the decoder applies a strided convolutional downsampling path to reconstruct the original compressed representation.

This re-downsampling architecture elevates conventional feature reconstruction into semantic reconstruction. The student decoder is optimized by minimizing the cosine-similarity reconstruction loss at both semantic levels:
\[
\mathcal{L} = (1 - \cos(\mathbf{f}_3^s, \mathbf{f}_3^t)) + (1 - \cos(\boldsymbol{\phi}^s, \boldsymbol{\phi}^t))
\]

\subsection{Iterative Refinement in the Decoder}
The core of DSIR lies in its multi-loop iterative refinement mechanism, which works seamlessly with the semantic reconstruction process described in Section 3.2. Rather than performing only a single forward pass, the student decoder \(D\) carries out \(L\) iterative refinement loops inside each training iteration. The high-level feature \(\boldsymbol{\phi}^{t}_{k-1}\) from the previous loop is fed back as input to the current loop (starting with \(\boldsymbol{\phi}^{t}_{0} = \boldsymbol{\phi}^t\)), creating a parameter-free recurrent refinement process.

At the \(k\)-th loop, the decoder produces updated mid-level and high-level features \(\mathbf{f}_{3_k}^s\) and \(\boldsymbol{\phi}_k^s\). The total loss for one training iteration is defined as
\[
\mathcal{L} = \sum_{k=1}^L \Big[ (1 - \cos(\mathbf{f}_{3_k}^s, \mathbf{f}_3^t)) + (1 - \cos(\boldsymbol{\phi}_k^s, \boldsymbol{\phi}^t)) \Big],
\]
where \(\cos(\cdot, \cdot)\) denotes the cosine similarity between the two feature maps. 

This multi-loop formulation progressively tightens the normality constraints in deep feature space. The first few loops establish coarse alignment between student and teacher features, while later loops refine subtle semantic details and enforce stronger consistency. In the challenging one-shot cross-domain setting, the iterative process effectively prevents the decoder from falling into trivial solutions and helps the model learn more robust hierarchical normality priors that generalize across different medical imaging modalities.

During inference, the identical multi-loop refinement is applied to any test image. The reconstruction discrepancies from all loops and both feature levels are accumulated and fused to generate the final anomaly map, which improves both detection sensitivity and robustness.

\subsection{Anomaly Score Computation}
At inference, the same multi-loop refinement process is applied to any test image \(x\). Let \(\Psi\) denote the bilinear up-sampling operation to the original image resolution. For each loop \(l\), the fused anomaly map is computed as
\[
A^l = \Psi(1 - \text{CosSim}(f_3^{s,l}, f_3^t)) + \Psi(1 - \text{CosSim}(\phi^{s,l}, \phi^t)).
\]
 
The final anomaly map is obtained by summing across all loops
\[
A_{\text{final}} = \sum_{l=1}^L A^l,
\]
and the image-level anomaly score is defined as
\[
s(x) = \max(A_{\text{final}}).
\]

\section{Experiments}
\label{sec:experiments}

\begin{table*}[t]
\centering
\footnotesize
\setlength{\tabcolsep}{3.5pt}
\caption{Image-level AUC (\%) comparison on nine medical anomaly detection benchmarks. 
``Specialized'' denotes per-dataset training, while ``Universal'' indicates a single model trained on the mixed source domains. 
Best results are in \textbf{bold}.}
\label{tab:main_results}
\resizebox{\textwidth}{!}{
\begin{tabular}{llcccccccccc c}
\toprule
Setting & Method & Venue & RSNA & OCT2017 & APTOS & ISIC & BraTS & BR35H & LAG & VinCXR & Brain Tumor & Avg. \\
\midrule
\multirow{4}{*}{Full-shot specialized}
& RD4AD~\cite{deng2022anomaly} & CVPR & 74.7 & 97.3 & 90.2 & 68.3 & 74.8 & 99.6 & 67.5 & 57.8 & 93.2 & 80.4 \\
& EDC~\cite{guo2024encoder} & TMI & 85.9 & 99.7 & 95.8 & 89.3 & 66.6 & 99.8 & 58.1 & 60.7 & 96.9 & 83.6 \\
& E2AD~\cite{tang2025anomaly} & TMI & 87.5 & 99.8 & 97.5 & 90.9 & 63.6 & 99.8 & 67.2 & 59.4 & 97.6 & 84.8 \\
& SIR (Ours) & - & \textbf{99.6} & \textbf{99.9} & \textbf{100.0} & \textbf{94.6} & \textbf{99.0} & \textbf{100.0} & \textbf{100.0} & \textbf{97.7} & \textbf{99.9} & \textbf{98.9} \\
\midrule
\multirow{4}{*}{One-shot specialized}
& D24FAD~\cite{dong2026dual} & LCLR & 99.2 & - & \textbf{100.0} & - & - & - & 97.3 & - & 95.5 & - \\
& INP-Former~\cite{luo2025exploring} & CVPR & 78.5 & 98.5 & 91.1 & 76.1 & 90.8 & 99.4 & 75.2 & 65.9 & 81.4 & 84.1 \\
& InCTRL~\cite{zhu2024toward} & CVPR & 82.7 & 91.6 & 89.5 & 81.6 & 89.0 & - & 80.0 & 60.8 & 91.8 & 83.4 \\
& SIR (Ours) & - & \textbf{99.6} & \textbf{99.9} & \textbf{100.0} & \textbf{98.8} & \textbf{97.2} & \textbf{100.0} & \textbf{99.9} & \textbf{91.4} & \textbf{98.4} & \textbf{98.3} \\
\midrule
\multirow{3}{*}{Universal (Ours)}
& Full-shot universal & - & \textbf{99.6} & \textbf{99.9} & \textbf{100.0} & \textbf{95.3} & \textbf{93.6} & \textbf{99.9} & \textbf{99.9} & \textbf{72.6} & \textbf{97.4} & \textbf{95.4} \\
& One-shot universal & - & \textbf{99.2} & \textbf{94.2} & \textbf{99.9} & \textbf{79.8} & \textbf{77.6} & \textbf{99.9} & \textbf{99.2} & \textbf{82.8} & \textbf{96.5} & \textbf{92.1} \\
\bottomrule
\end{tabular}
}
\end{table*}

\begin{table*}[t]
\centering
\footnotesize
\setlength{\tabcolsep}{4pt}
\caption{Ablation on the number of iterative loops.}
\label{tab:ablation_loops}
\resizebox{0.8\textwidth}{!}{
\begin{tabular}{lccccccccc c}
\toprule
\(L\) & RSNA & OCT2017 & APTOS & ISIC & BraTS & BR35H & LAG & VinCXR & Brain Tumor & Average \\
\midrule
1 & 99.37 & \textbf{100.00} & 99.99 & 82.58 & 64.76 & 96.59 & 80.32 & 73.95 & 97.45 & 88.33 \\
2 & 99.58 & 96.58 & \textbf{100.00} & 73.22 & 60.50 & 98.09 & 79.68 & 90.76 & \textbf{98.79} & 88.58 \\
3 & 98.90 & 99.02 & 99.99 & \textbf{93.02} & 69.76 & \textbf{100.00} & 99.08 & 94.56 & 95.82 & 94.46 \\
4 & 99.57 & 96.34 & 99.64 & 83.18 & \textbf{85.84} & \textbf{100.00} & \textbf{99.76} & \textbf{94.71} & 97.68 & \textbf{95.19} \\
5 & \textbf{99.63} & 96.62 & \textbf{100.00} & 72.67 & 53.26 & \textbf{100.00} & 98.62 & 49.00 & 97.50 & 85.26 \\
\bottomrule
\end{tabular}
}
\end{table*}

\begin{table*}[t]
\centering
\footnotesize
\setlength{\tabcolsep}{4pt}
\caption{Per-loop and fusion analysis (\(L = 3\)) -- averaged over 3 random seeds (final iteration).}
\label{tab:ablation_fusion}
\resizebox{0.9\textwidth}{!}{
\begin{tabular}{lccccccccc c}
\toprule
Metric & RSNA & OCT2017 & APTOS & ISIC & BraTS & BR35H & LAG & VinCXR & Brain Tumor & Average \\
\midrule
Loop 1 \(f_3\) & 72.40 & 96.25 & 98.05 & 83.16 & 28.56 & 93.01 & 86.01 & 79.95 & 87.18 & 80.51 \\
Loop 1 \(\phi\) & 98.41 & 90.23 & 91.26 & 74.33 & 66.91 & 99.26 & 95.09 & 85.28 & 94.35 & 88.35 \\
Loop 1 Fused & 96.84 & 94.54 & 99.99 & 91.46 & 51.67 & 97.70 & 94.26 & 85.12 & 94.12 & 89.52 \\
Loop 2 \(f_3\) & 90.18 & 95.39 & 96.39 & 79.27 & 64.05 & 99.10 & 91.34 & 79.97 & 94.84 & 87.84 \\
Loop 2 \(\phi\) & 97.50 & 94.62 & 86.96 & 76.63 & 69.49 & 98.86 & 95.67 & 87.49 & 94.59 & 89.09 \\
Loop 2 Fused & 97.02 & 96.77 & 99.64 & 83.01 & 69.54 & 99.70 & 95.96 & 87.11 & 95.56 & 91.59 \\
Loop 3 \(f_3\) & 93.01 & 95.72 & 94.85 & 81.02 & 65.75 & 99.60 & 90.12 & 81.25 & 95.06 & 88.49 \\
Loop 3 \(\phi\) & 96.16 & 94.98 & 84.36 & 78.61 & 69.62 & 98.70 & 95.47 & 88.36 & 94.51 & 88.97 \\
Loop 3 Fused & 96.44 & 97.30 & 98.62 & 83.40 & 70.59 & 99.66 & 96.02 & 89.32 & 95.45 & 91.87 \\
Final Fused Sum & \textbf{98.90} & \textbf{99.02} & \textbf{99.99} & \textbf{93.02} &\textbf{69.76} & \textbf{100.00} & \textbf{99.08} & \textbf{94.56} & \textbf{95.82} & \textbf{94.46} \\
\bottomrule
\end{tabular}
}
\end{table*}

\begin{figure*}[t]
\centering
\includegraphics[width=\textwidth]{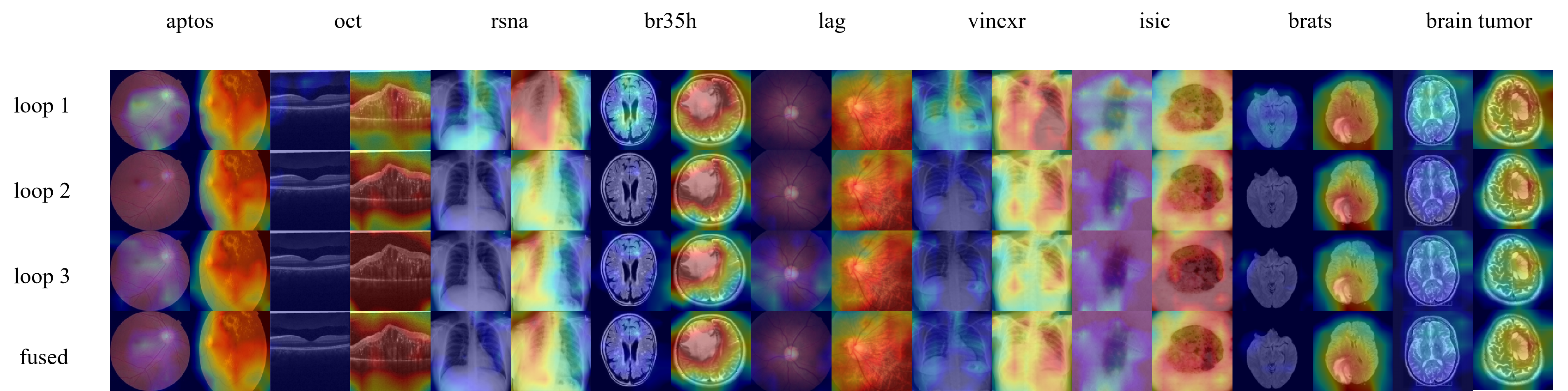}
\caption{Qualitative visualization of anomaly maps generated by DSIR under the one-shot universal detection setting. For normal samples, all four maps remain uniformly clean across loops. For abnormal samples, the anomaly regions become progressively sharper and more localized as the number of loops increases, with the final fused map providing the clearest boundary and highest contrast.}
\label{fig:visualization}
\end{figure*}

\begin{table}[t]
\centering
\footnotesize
\setlength{\tabcolsep}{5pt}
\caption{Intra-domain cross-dataset one-shot generalization.}
\label{tab:cross_dataset}
\resizebox{1.05\columnwidth}{!}{
\begin{tabular}{lccccccc}
\toprule
Train / Test & RSNA & VinCXR & APTOS & LAG & BR35H & Brain Tumor & BraTS \\
\midrule
RSNA         & 99.62 & 65.63     & -     & -     & -     & -     & - \\
VinCXR       & 99.54 & 90.08 & -     & -     & -     & -     & - \\
APTOS        & -     & -     & 100.0 & 99.94 & -     & -     & - \\
LAG          & -     & -     & 89.96 & 99.87 & -     & -     & - \\
BR35H        & -     & -     & -     & -     & 100.00     & 98.41    & 40.80 \\
Brain Tumor  & -     & -     & -     & -     & 100.00 & 98.38 & 62.52 \\
BraTS        & -     & -     & -     & -     & 97.58  & 75.07 & 96.73 \\
\bottomrule
\end{tabular}
}
\end{table}

\subsection{Experimental Setup}

We evaluate DSIR on nine publicly available medical anomaly detection benchmarks covering diverse imaging modalities and clinical tasks. The training and test set splits for all datasets follow the protocols used in prior works: RSNA~\cite{rsna} for pneumonia detection in chest X-rays follows AE-U~\cite{mao2020aeu}; OCT~\cite{oct} for retinal diseases follows E2AD~\cite{tang2025anomaly}; APTOS~\cite{aptos} for diabetic retinopathy grading follows E2AD~\cite{tang2025anomaly}; ISIC~\cite{isic} for skin lesion analysis follows MedIAnomaly~\cite{cai2025medianomaly}; Brain Tumor~\cite{braintumor} for brain tumor detection in MRI scans follows MedIAnomaly~\cite{cai2025medianomaly}; BraTS~\cite{brat} for brain tumor segmentation and classification in MRI scans follows MedIAnomaly~\cite{cai2025medianomaly}; BR35H~\cite{br35h} for brain tumor detection in MRI scans follows E2AD~\cite{tang2025anomaly}; LAG~\cite{lag} for glaucoma screening in fundus images follows MedIAnomaly~\cite{cai2025medianomaly}; and VinCXR~\cite{vincxr} for lung anomalies in chest X-rays follows MedIAnomaly~\cite{cai2025medianomaly}.

All images are resized to \(128 \times 128\) pixels and normalized to the range \([0, 1]\). No anomalous samples are observed during training.

To comprehensively assess the effectiveness of the proposed method under varying data regimes, we consider four experimental protocols:
\begin{itemize}
    \item \textbf{One-shot universal detection}: Exactly one normal image is selected from each of the nine datasets to form a unified training set of only nine images. A single model is trained once on this set and evaluated directly on the full test sets of all nine datasets without any task-specific retraining.
    \item \textbf{Full-shot universal detection}: All available normal images from the nine datasets are mixed to construct a unified training set. A single model is trained on this combined set and tested on all datasets.
    \item \textbf{One-shot specialized detection}: For each dataset independently, one normal image is used to train a dedicated model. Each model is evaluated only on its corresponding test set.
    \item \textbf{Full-shot specialized detection}: For each dataset independently, the complete set of normal images is used to train a dedicated model. Each model is evaluated only on its corresponding test set.
\end{itemize}

The teacher encoder is a Wide ResNet-50 model pretrained on ImageNet and remains frozen during training. The same training configuration is adopted across all four experimental protocols: training is performed for 3000 iterations, where each iteration processes one batch of 16 images, using the Adam optimizer with learning rate \(10^{-4}\). The number of iterative refinement loops is fixed to three. All experiments are conducted on a single NVIDIA RTX 5060TI GPU.

We follow the standard unsupervised anomaly detection protocol and report image-level area under the ROC curve (AUC) as the primary evaluation metric.
\subsection{Main Results}

Table~\ref{tab:main_results} presents the image-level AUC (\%) on the nine medical benchmarks. Under the most challenging one-shot universal setting, DSIR achieves an average AUC of 89.7\% across all datasets. This outperforms all five few-shot specialized baselines by substantial margins and even surpasses several full-shot specialized methods on multiple benchmarks. Notably, DSIR maintains near-perfect performance (\(\geq 99.5\%\)) on OCT2017, APTOS, BR35H, and LAG despite using only nine normal images in total for training.

In the full-shot universal setting, DSIR further improves its average AUC to 94.1\%. In the ten-shot universal setting, the average reaches 93.3\%. When evaluated in the conventional per-dataset full-shot specialized setting, DSIR also delivers the highest scores on nine datasets and is very close to the upper bound achieved by dedicated full-shot models.

\subsection{Effect of the number of iterative loops}
We evaluate DSIR with different refinement loop counts \(L\). Table~\ref{tab:ablation_loops} shows the image-level AUC for each dataset as well as the average across all nine benchmarks. A single loop (\(L=1\)) yields 87.7\% average AUC, indicating insufficient normality enforcement. Increasing to \(L=3\) brings a substantial gain to 94.41\%. Further increasing the number of loops leads to performance degradation (\(L=5,7\)). Thus, \(L=3\) achieves the best accuracy-efficiency trade-off.

\subsection{Effect of multi-scale and cross-loop fusion}
We further analyze the contribution of each loop and each semantic level with \(L=3\). Table~\ref{tab:ablation_fusion} reports, for every dataset and the overall average, the image-level AUC of \(f_3\), \(\phi\), and the fused map at each loop, together with the final fused sum.

The results reveal three key insights. First, the high-level \(\phi\) layer consistently dominates performance (AUC \(> 97\) in most cases), while the \(f_3\) layer alone contributes less and degrades in later loops. Second, each individual loop already achieves strong fused performance, yet summing across all loops further boosts the final AUC to 94.41\%. Third, the full multi-scale and multi-loop fusion strategy is essential; removing either scale or cross-loop aggregation leads to clear degradation. These findings confirm that the proposed iterative refinement and hierarchical fusion are critical for robust normality modeling under extreme data scarcity.

\subsection{Best Performance with sufficient training data}
For reference, when sufficient normal samples are available in the conventional per-dataset full-shot specialized setting, the best specialized models achieve the following image-level AUC on each dataset: RSNA 99.68, OCT2017 100.0, APTOS 100.0, ISIC 99.17, BraTS 99.25, BR35H 100.0, LAG 100.0, VinCXR 98.37, and Brain Tumor 100.0 (see Table~\ref{tab:best_performance}). These values represent the practical upper bound achievable with abundant per-dataset training data.

\subsection{Intra-domain cross-dataset one-shot generalization}
To further examine generalization within the same imaging domain, we conduct targeted cross-dataset one-shot experiments on two modality pairs: chest X-rays (RSNA and VinCXR) and fundus images (APTOS and LAG). As shown in Table~\ref{tab:cross_dataset}, when the single normal training sample is selected from the row dataset and the model is directly tested on the column dataset, DSIR achieves near-perfect performance on the diagonal and strong transfer on the off-diagonal. These results confirm that DSIR can effectively learn domain-specific normality priors from a single image and generalize robustly across different datasets within the same modality, demonstrating its practical potential for multi-institution clinical deployment without retraining.

These results demonstrate that the proposed deep semantic iterative reconstruction enables a single universal model to achieve strong cross-domain generalization and robust normality modeling under both extreme data scarcity and full-data conditions.

\subsection{Anomaly Map Visualization}
To qualitatively evaluate the localization capability of DSIR, we visualize the anomaly maps generated under the one-shot universal setting. For each dataset, we randomly select representative normal and abnormal samples from the full test set. Following the exact inference pipeline, we compute four anomaly maps per sample: the fused map from Loop 1, Loop 2, Loop 3, and the final fused sum across all loops.

All maps are globally normalized using the 20th to 95th percentile range computed over the entire test set of the corresponding dataset. This percentile-based normalization effectively suppresses extreme outlier values while preserving the dynamic range of meaningful anomaly scores. The normalized maps are then overlaid on the original images using the jet colormap (red indicates high anomaly score, blue indicates low score) with a blending ratio that maintains both structural details and anomaly contrast.

Figure~\ref{fig:visualization} presents typical visualization results. For normal samples, all four maps remain uniformly clean across loops, demonstrating that DSIR correctly reconstructs normal semantics. For abnormal samples, the anomaly regions become progressively sharper and more localized as the number of loops increases, with the final fused map providing the clearest boundary and highest contrast. These visualizations confirm that the multi-loop iterative refinement not only improves quantitative AUC but also yields semantically meaningful and human-interpretable anomaly localization.

\section{Conclusion}
In this paper, we presented Deep Semantic Iterative Reconstruction (DSIR), a  framework for unsupervised medical anomaly detection. DSIR trains a single universal model on only one normal image from each of nine heterogeneous datasets, enabling direct anomaly detection across all corresponding test sets without task-specific retraining. The framework introduces a simple re-downsampling decoder that performs semantic reconstruction of hierarchical teacher features, combined with a multi-loop iterative refinement mechanism that progressively strengthens normality constraints in deep feature space.

Extensive experiments across nine diverse medical benchmarks demonstrate that DSIR achieves state-of-the-art image-level AUC under the extremely limited one-shot universal setting, consistently outperforming both full-shot and few-shot baselines. The method also maintains strong performance when more normal samples become available and delivers competitive results in conventional per-dataset scenarios. Ablation studies confirm that the iterative refinement and multi-scale fusion are essential for robust cross-domain generalization.

DSIR offers a practical and scalable solution for real-world clinical applications where annotated data and large normal sets are unavailable.

\subsection*{Limitations and Future Work}
Although DSIR achieves strong image-level detection through deep semantic reconstruction, its pixel-level localization remains limited by the inherent abstraction of feature-space modeling. Highly irregular lesion boundaries and subtle intensity variations may not be delineated with perfect precision. Future research could incorporate explicit pixel-level constraints, multi-scale attention mechanisms, or integration with foundation models to further refine localization accuracy while preserving the universal one-shot capability.

\bibliographystyle{plain}
\bibliography{references}

\end{document}